\documentclass[sigconf,table]{acmart}

\setcopyright{rightsretained}
\acmDOI{10.475/123_4}

\acmISBN{123-4567-24-567/08/06}

\acmConference[LONDON'18]{ACM SigKDD conference}{July 2018}{London, U.K.}
\acmYear{2018}
\copyrightyear{2018}

\acmArticle{4}
\acmPrice{15.00}

\usepackage[utf8]{inputenc} % allow utf-8 input
\usepackage[T1]{fontenc}    % use 8-bit T1 fonts
\usepackage{hyperref}       % hyperlinks
\usepackage{url}            % simple URL typesetting
\usepackage{booktabs}       % professional-quality tables
\usepackage{amsfonts}       % blackboard math symbols
\usepackage{nicefrac}       % compact symbols for 1/2, etc.
\usepackage{microtype}      % microtypography
\usepackage{changepage}
\usepackage{natbib}

\usepackage[makeroom]{cancel}
\usepackage{amssymb}
\usepackage{dsfont}
\usepackage{amstext}
\usepackage{multirow}
\usepackage{amsmath}

\usepackage{algorithm}
\usepackage{multirow}
\usepackage{graphicx}
\usepackage{subfig}
\usepackage{todonotes}

\usepackage{algcompatible}

\algblockdefx{FORALLP}{ENDFAP}[1]%
  {\textbf{for }#1 \textbf{do in parallel}}%
  {\textbf{end for}}
  
\usepackage{threeparttable,booktabs}
\usepackage{array}
\newcolumntype{@}{>{\global\let\currentrowstyle\relax}}
\newcolumntype{^}{>{\currentrowstyle}}
\newcommand{\rowstyle}[1]{\gdef\currentrowstyle{#1}%
  #1\ignorespaces
}
\definecolor{hcolor1}{rgb}{0.90,0.90,0.90}
\definecolor{hcolor2}{rgb}{0.94,0.94,0.94}
\definecolor{hcolor3}{rgb}{0.98,0.98,0.98}

\begin{document}
\title{Heron Inference for Bayesian Graphical Models}

%\titlenote{Produces the permission block, and copyright information}
%\subtitle{Extended Abstract}
%\subtitlenote{The full version of the author's guide is available as \texttt{acmart.pdf} document}

\author{Daniel Rugeles}
\orcid{0000-0001-8085-3123}
\affiliation{%
  \institution{Nanyang Technological University}
  \country{Singapore}
  \postcode{43017-6221}
}
\email{daniel007@e.ntu.edu.sg}

\author{Manoranjan Dash}

\affiliation{%
  \institution{National University of Singapore}
  \country{Singapore}
}
\email{mano@comp.nus.edu.sg}

\author{Zhen Hai}

\affiliation{%
  \institution{Institute for Infocomm Research}
  \country{Singapore}}
\email{haiz0001@ntu.edu.sg}

\author{Gao Cong}
\affiliation{%
  \institution{Nanyang Technological University}
  \country{Singapore}
}
\email{gaocong@ntu.edu.sg}

\begin{abstract}
Bayesian graphical models have been shown to be a powerful tool for discovering uncertainty and causal structure from real-world data in many application fields. Current inference methods primarily follow different kinds of trade-offs between computational complexity and predictive accuracy.  At one end of the spectrum, variational inference approaches perform well in computational efficiency, while at the other end, Gibbs sampling approaches are known to be relatively accurate for prediction in practice. In this paper, we extend an existing Gibbs sampling method, and propose a new deterministic Heron inference (Heron) for a family of Bayesian graphical models. 
In addition to the support for nontrivial distributability, one more benefit of Heron is that it is able to not only allow us to easily assess the convergence status but also largely improve the running efficiency.
We evaluate Heron against the standard collapsed Gibbs sampler 
and state-of-the-art state augmentation method
in inference for well-known graphical models.       
Experimental results using publicly available real-life data have demonstrated that
Heron significantly outperforms the baseline methods for inferring Bayesian graphical models.

%In this paper we propose \textit{Heron Inference}, a deterministic method that yields higher  efficiency and predictive accuracy in addition to trivial distributability. The method can be applied to models where the latent variables are independent given the data and other parameters. We evaluate our work by comparing the predictive power, time efficiency and distributability of Heron Inference with the standard collapsed Gibbs sampler and an state of the art inference method within the Gibbs framework. We apply the inference methods to three topic models facilitating the assessment on the extensibility of the inference method to other models. Results show improved perplexity and running time on different batch sizes for three publicly available datasets.% We assess the predictive power and time efficiency of our proposed inference method on three datasets using the Relational Topic Model, Latent Dirichlet Allocation and Supervised Latent Dirichlet Allocation.
% * <danrugeles@gmail.com> 2017-09-07T04:55:31.143Z:
%
% ^.
% * <danrugeles@gmail.com> 2017-09-07T04:55:30.773Z:
%
% ^.
\end{abstract}

\maketitle

\section{Introduction}

Bayesian graphical models refer to a family of probabilistic unifying models in which nodes represent random variables, and edges represent possible dependency between individual pairs of nodes. The graphical models have become a powerful tool for discovering uncertainty and causal structure from real-world data. They have been widely used in computer vision \cite{cvtopic}, text data mining \cite{LDA}, \cite{RTM}, and \cite{sLDA}, natural language processing \cite{jointtopic}, music information retrieval \cite{musictopic}, and computational biology \cite{genotopic}. 
Given a graphical model, one of the most fundamental tasks could be \textit{inference}, simply, computing the marginal distribution of one or a couple of random variables in the model.

Generally, two categories of approaches are mainly adopted for inferring Bayesian graphical models, i.e., variational inference and Gibbs sampling methods. Variational approaches mainly rely on  stochastic optimization to fit an approximate model out of a postulated family of models. The choice for the family is known to facilitate efficiency as well as online learning process of the algorithms \cite{ovlda}. However, the approximation for the targeted true model via a different model inevitably sacrifices accuracy. In contrast, based on the theory of stationary distribution that guarantees a convergence to the desired distributions, Gibbs sampling methods have been shown to result in decent predictive accuracy. Unfortunately, their convergence is often costly and difficult to assess. The Gibbs sampling methods are typically slow compared to variational inference, due to the fact that the posterior distribution must be computed and sampled for every single observation. Moreover, both collapsed Gibbs sampling and variational inference are known to be sequential algorithms, meaning that every step in the methods depends on the results of the previous step. Then, these algorithms are basically not distributable.

Recently, a state augmentation inference method, called state augmentation for marginal estimation (SAME)\cite{cooled}, has shown that by drawing a number of samples as opposed to extracting one sample from each conditional distribution, it is possible to improve the predictive perplexity of the Gibbs sampler and reduce the negative impact on perplexity caused by distributing the datasets into batches. Therefore, a higher sampling replication rate is able to yield better predictive performance. Unfortunately, SAME's computational complexity depends on the number of replications. As the number of replications increases, its computational cost will eventually surpass the computational cost of the Gibbs sampler. To keep the computational cost reasonable, SAME  typically performs $100$ replications for sampling from the conditional distribution. Arguably, SAME could be the fastest and most accurate inference algorithm under the Gibbs sampling framework. 

In this paper, we aim at improving the computational cost and predictive power in inference for Bayesian graphical models. 
We extend SAME, and develop a new Heron inference approach.
In particular, we develop a new learning framework that drives the number of replications of latent state up to infinity. By doing so, we then yield a deterministic inference algorithm even within the Gibbs framework. Determinism is typically a major source for computational speed-up. Though modern GPUs support hardware implementations of some samplers, sampling still limits the performance of randomized inference algorithms. Our proposed heron Inference method does not require sampling, which leads to much better efficiency than sampling counterparts. In addition, maximizing the number of replications of the sampling procedure naturally leads to an improvement for the predictive power of the algorithm. In addition, the easy assessment for convergence is possible thanks to the determinism. %More on this in later sections.

With analyzing the convergence properties, we found that the proposed deterministic inference method is ultimately solving a fixed-point system of equations. This is perhaps one of the major benefits of Heron inference, as fixed-point methods have been well studied, and their iterative solvers can be clearly distributable. An additional advantage from the fixed-point system is that the computation of repeated document-word tuples is unnecessary, which actually leads to an improvement for the efficiency. Although recently proposed inference algorithms also avoid this cost \cite{ldacvb}, the fixed-point system provides a theoretical explanation within the Gibbs framework. 

%% MAYBE MENTION ON CONVERGENCE!

 % As such, we present Heron inference, a novel distributable inference method more accurate and faster than cooled inference method. We maximize the replication of samples used by Cooled to infinity and we find that collapsed Gibbs sampler becomes a deterministic algorithm where the inference of the parameters can be extracted from the solution to a system of non linear equations. To this end, we organize this document as follows: 

%We use the recent findings of deterministic Gibbs motion (DGM) [dgm], a deterministic inference method that uses the information of the whole posterior -- and not just the sampling of one dimension -- to update the statistics of the model. DGM is more accurate than collapsed Gibbs Sampling while maintaining the computational efficiency. In addition, the deterministic nature of DGM allow for the use of new equations from which we observe convergence properties that can be exploited to transform the inference of LDA into solving a system of non-linear equations. We apply the fix-point method to solve this system and we show that the solution converges to the solution of DGM which yields three benefits 1) The algorithm is distributable, 2) The computational complexity does not depend on the number of words that appear in all document. Instead, it depends on the number of distinct number of words that appears on each document and 3) less expensive and easier convergence assessment.

The proposed Heron inference can be applied to a family of graphical models that satisfy the following conditions: 1) The posterior distributions of the model need to have an analytical form, 2) The posterior distributions need to be discrete distributions,, and 3) The latent random variables need to be independent of each other, given the data and related parameters of the model. Note that we coin the term \textit{Heron Inference} in honor of Heron of Alexandria. Heron is perhaps thought to be the first person who solved iteratively a fixed-point function with the objective of computing the square root of a number. %Without loss of generality, we follow the analysis of our method applied to the case of the Latent Dirichlet Allocation \cite{LDA}. However we also apply our method to the relational Topic model and the supervised Latent Dirichlet Allocation as presented in the experimental section.

We have made the following main contributions in this work:
\begin{enumerate}
\item We propose a novel inference method, which is applicable to a family of probabilistic models, with the same convergence guarantees as Gibbs sampling but easier convergence assessment.
\item Our proposed method is an order of magnitude faster method than the state-of-the-art inference method available within the Gibbs framework, and at the same time achieving an improved perplexity.
\item We demonstrate theoretical support for the distributability of the proposed algorithm.
\item We transform the Gibbs sampling inference approach into a new deterministic domain where optimization techniques can be further explored.
\end{enumerate}

\section{Related Work}

%Introduction
Along rich history of Bayesian graphical models, several main categories of inference algorithms have been developed, i.e., variational Bayesian inference \cite{LDA}, expectation propagation \cite{ldaep}, Gibbs sampling \cite{ldacgs}, and belief propagation \cite{ldabp}. Among these, Gibbs sampling  and variational inference are perhaps the most popular algorithms, possibly due to their efficacy or efficiency.  Specifically, Gibbs sampling methods have been known to guarantee to converge to the true posterior via a sampling scheme, while the variational inference basically relies on a theoretically-backed optimization approach to approximate the true posterior.% As such the sampling schemes target an improvement in accuracy as typically measured by the predictive perplexity, while the optimization schemes target 

%% Improvements of VI
Existing approaches in the category of variational inference aim to improve the predictive perplexity. These methods approximate intractable integrals, which then results in inaccuracy. Collapsed variational Bayesian schemes \cite{ldacvb} use a second-order Taylor expansion to approximate the integrals, while other variational approaches use the zero-order information in order to obtain more accurate inference \cite{cvb0}. With respect to the distributability of these methods, empirical work has demonstrated effective approximations that avoid a major impact on the predictable performance \cite{ovlda}.

%% Improvements of Collapsed Gibbs Sampling
% Many improvements have been made to LDA-CGS. Given the predictive power of this method, most of the research is focused on improving its speed while sacrificing little to zero predictive power. 
Previous methods in the category of Gibbs sampling are mainly concerned with improving efficiency. This is natural given their powerful predictive capability. FastLDA was shown to be $8$ times faster than the latent Dirichlet allocation via collapsed Gibbs sampling (LDA-CGS), while maintaining the exact same predictive power \cite{fastlda}. The high computational cost of LDA-CGS stems from the $O(K)$ time complexity incurred at every sampling step (K: number of latent topics). One key observation is that many words are often assigned to only a small number of different topics. Hence, by keeping track of these words, FastLDA requires significantly less than $K$ operations per sample on average. 

SparseLDA \cite{sparsefastlda} factorizes the posterior equation into a sum of three factors. The sampling scheme is then replaced by a uniform sampling, where it is observed that the probability mass falls in one of the buckets 90\% of the time. By making an appropriate data structure, the computation of the mass can be optimized. As a result, the model achieves an order of magnitude improvement in computational complexity without affecting its predictive power, when compared to collapsed Gibbs sampling. A faster approach, which exploits sparsity, relis on a variant of the Fenwick tree to encode the posterior, so that sampling can be performed in $log(K)$ time (F+LDA) \cite{F+LDA}.

Based on efficient Metropolis-Hastings, AliasLDA \cite{aliaslda} and LightLDA \cite{lightlda} can reduce the $O(K)$ complexity of sampling to $O(1)$ via Walker's alias method \cite{aliastable}. %As explained by the authors: \textit{At its core is the idea that dense, slowly changing distributions can be approximated efficiently by the combination of a Metropolis-Hastings step, use of sparsity, and amortized constant time sampling via Walker's alias method}. 
Unfortunately, these methods result in problems of frequent cache misses caused by random accesses to the parameter matrices. WarpLDA deals with the Metropolis-Hasting's problem by fitting the use of randomly accessed memory per-document in the $L3$ cache. Remarkably, WarpLDA is consistently $5 \sim 15$ times faster than LightLDA, and it also outperforms F+LDA \cite{WarpLDA}.

%% GPU Implementations
The sparsity exploited by recent algorithms has been used to develop GPU implementations of inference algorithms \cite{saberLDA}. Alternatively, the SAME method is the inference algorithm that leverages a Poisson distribution and efficiently replaces sampling several times a categorical distribution with sampling a single time a Poisson distribution \cite{cooled}. Their GPU implementation is much faster than the GPU algorithm first introduced in \cite{GLDA}. Different from other GPU based methods, Poisson sampling is done via a hardware implementation available on \textit{NVIDIA} GPUs, which makes SAME a very efficient inference algorithm.

Existing inference methods reside in the Pareto optima, where variational approaches are optimal in efficiency, while sampling approaches are optimal in predictive power. %as measured by the predictive perplexity. 
Other inference methods for Bayesian graphical models have different trade-offs between efficiency and accuracy. Recently, SAME has been shown to improve the optima using the state augmentation technique. The resulting algorithm is more accurate and efficient than existing CPU or GPU implementations \cite{cooled}. In this work, we propose to maximize the augmentation of the latent state, and then develop a deterministic algorithm whose convergence properties can be exploited to further improve the efficiency and efficacy for inferring graphical models in practice.

%% Deep Learning approaches to LDA
%The inference of topic models goes beyond the field of probability, deep learning models have been proposed to approximate the learning of the parameters of LDA. The Neural Topic Model proposes to fit the likelihood function of LDA in a network where the weights are normalized to fit probabilities \cite{NTM}. However, there are no theoretical links to the topic model.

%In the recent years, numerous deep learning frameworks have been developed. Deep learning frameworks are attractive partly because they allow rapid development of new models as well as automatic execution of the code in GPU or distributed settings. Several efforts have been made to port probabilistic models into . However, most of them are follow variational inference approaches given the fact that both VI inference techniques and deep learning models are stated as an optimization problem. In some cases, Deep Learning follow the intuition of probabilistic models so as to . However, this is not scalable because every model end up being a per model basis approximation models For example, et al \cite{NTM}. 

% The contribution in this paper is rather of an odd one. We contribute to LDA-CGS by transforming the sampling scheme into a deterministic scheme which results in improving its predictive power while maintaining its time complexity.

\section{Preliminaries}

%\begin{table*}[t]
%  \caption{Statistics used to compute the posterior of LDA, RTM, and sLDA}
%  \label{tab:equations}
%  \centering
%  \begin{tabular}{llll}
%    \toprule
%    Model    & Hyperparameters & Statistics $\Omega$ & Unnormalized posterior density $g(\Omega)$    \\
%    \midrule
%    LDA & $\alpha$, $\beta$ & $C^{-j}_{k_j,d_j}$, $D^{-j}_{k_j,w_j}$, $D^{-j}_{k_j,.}$ &$(C^{-j}_{k_j,d_j}+\alpha) %\dfrac{D^{-j}_{k_j,w_j}+\beta}{D^{-j}_{k_j,.}+V\beta}$      \\
%    RTM  &$\alpha$, $\beta$, $b$  & $C^{-j}_{k_j,d_j}$, $D^{-j}_{k_j,w_j}$, $D^{-j}_{k_j,.}$, $n_{d',k}$ %&$(C^{-j}_{k_j,d_j}+\alpha) \dfrac{D^{-j}_{k_j,w_j}+\beta}{D^{-j}_{k_j,.}+V\beta}exp\left(\dfrac{b_k}{N_d}\sum %\dfrac{n_{d',k}}{N_{d'}}\right)$      \\
%    sLDA  & $\alpha$, $\beta$, $b$, $a$ & $C^{-j}_{k_j,d_j}$, $D^{-j}_{k_j,w_j}$, $D^{-j}_{k_j,.}$,  & %$(C^{-j}_{k_j,d_j}+\alpha) \dfrac{D^{-j}_{k_j,w_j}+\beta}{D^{-j}_{k_j,.}+V\beta}exp\left(-(y_d-b^T\bar{z}_d-a)^2\right)$   %\\
%    \bottomrule
%  \end{tabular}
%\end{table*}

In this section, 
we present preliminaries about existing well-known graphical models and the unified collapsed Gibbs sampling algorithm, followed by the state-of-the-art state augmentation inference method.

Latent Dirichlet allocation (LDA), one of the most popular graphical models, has been widely used to discover latent topical structure of a given text collection of documents.
%Latent Dirichlet Allocation (LDA) is a generative model for a collection of $|\mathbb{D}|$ documents and its %set of words $w_{di}$ for the $i^{th}$ position in the $d^{th}$ document. 
Simply, LDA assumes the following generative process:

\begin{center}
$\theta_d \sim Dirichlet(\alpha)$ ~~
$\phi \sim Dirichlet(\beta)$
$z_{di} \sim Discrete(\theta_d)$ ~~
$w_{di} \sim Discrete(\Phi_{z_{di}})$
\end{center}
where $\theta_d$ corresponds to per-document topic distribution of document $d$ ($d: 1 \sim |\mathbb{D}|$), $\phi_k$ corresponds to per-topic word distributions for topic $k$ ($k: 1 \sim K$), $z_{di}$ corresponds to latent topic assignment for word $w_{di}$ ($i: 1 \sim N_d$, number of word tokens in document $d$), 
and $\alpha$ and $\beta$ are hyper-parameters.

%The inference of LDA is typically performed using MCMC approaches [cgs] or Optimization approaches [vb]. We use the theory behind MCMC to obtain an algorithm with higher accuracy and less computational complexity than the standard collapsed Gibbs sampling (CGS). In addition, we obtain an algorithm that is trivially distributable.

Relational topic model (RTM) \cite{RTM} extends LDA by taking into account the citation links between pairwise documents. RTM adds one more step to the generative process of LDA:

\begin{center}
$y_{d,d'} \sim \psi(.|\bar{z_{d}},\bar{z_{d'}})$ 
\end{center}
where $y_{d,d'}$ is an indicator variable of the citation relationship between documents $d$ and $d'$, and
its probability is given by the following function:

\[\psi(y_{d,d'}{=}1)=exp\left(\eta^{T}(\bar{z_d} \odot \bar{z_d'}+ \nu)\right)\]
where $\eta$ and $\nu$ are hyper-parameters, $\bar{z_d}=\dfrac{1}{N_d}\displaystyle\sum_{i{=}1}^{N_d}z_{d,i}$ ($z_{d,i}$: one-hot based topic assignment), and $\odot$ means element-wise product.

Supervised LDA (sLDA) \cite{sLDA} is a statistical model of labeled documents, and it
 extends LDA by adding an observed variable $x$ that ndicates the label of a document,
%as specified by the training data. 
for example, the rating given to a movie. 
The sLDA adds the following step to the generative process of LDA:

\begin{equation}
\label{eq:super}
x_{d} \sim N(\gamma^{T}\bar{z_{d}},\sigma^2) 
\end{equation}
where $x_d$ is the label of document $d$, 
%corresponds to an unconstrained ordered value 
and both $\gamma$ and $\sigma$ are hyper-parameters. 
%Note that different functions can be proposed to fit how the topic assignments generate the supervised %variable $x$. However, we apply Equation \ref{eq:super} to the experiments presented in Section %\ref{sec:Experiments}.

\begin{algorithm}[t]
   \caption{Collapsed Gibbs Sampling (CGS)}
   \label{alg:cgs}
\begin{algorithmic}[1]
   \STATE {\bfseries Input:} Dataset $\mathbb{D}$ (size: $|\mathbb{D}|$), latent assignment $z_j$
   \STATE \textbf{Initialize} $z_j, \forall z_j \in 1 \cdots  |\mathbb{D}|$, $\Omega$ = computeStats($z_j$)
   %$\Omega$ = $\{C, D\}$
   \REPEAT
   \FOR{$j=1$ {\bfseries to} $|\mathbb{D}|$}
   \STATE updateStats($\Omega,z_j$, $Substract$)
   \STATE  $p(Z_j|Z_{-j}={z_{-j}},\mathbb{D}) \propto g(\Omega)$ from Table \ref{tab:equations_cgs}
   \STATE $z_j'$=sample($p(Z_j|Z_{-j}={z_{-j}},\mathbb{D})$)
   \STATE updateStats($\Omega,z_j'$, $Add$) from Equation \ref{eq:updatestats}
   \ENDFOR
   \UNTIL{convergence}
\end{algorithmic}
\end{algorithm}

The inference for these graphical models, %along with other models such as the Stochastic Block Mixture Model \cite{SBMM} 
can be typically performed using collapsed Gibbs sampling (CGS). 
Algorithm \ref{alg:cgs} describes the unified sampling method by sequentially processing each of the $|\mathbb{D}|$ records in the dataset.
Table \ref{tab:equations_cgs} lists the posterior density of each graphical model 
as a function of the statistics $\Omega$. 
Some main statistics are defined as follow:
%(computeStats in Algorithm \ref{alg:cgs}):
$C_{k_j,d_j}$: the number of times that assignment $k_j$ and document $d_i$ co-occur in the training data. 
%$C^{-j}_{k_j,d_j}$ without considering the $j^{th}$ record, 
 $D_{k_j,w_j}$: the number of counts that latent assignment $k_j$ and word $w_i$ co-occur in the data.
% $D^{-j}_{k_j,w_j}$ without considering the $j^{th}$ record, 
$D_{k_j,.}=\sum_w D_{k_j,w_j}$: the number of counts that latent assignment k occurs in the given data. 
%$D^{-j}_{k_j,.}=\sum_w\phi^{-j}_{k_j,w_j}$ without considering the $j^{th}$ record.
Note that ``-j'' in a quantity means that the contribution of the $j^{th}$ record is not considered in the quantity.

\begin{table}[h]
  \caption{Posterior density as a function of the statistics $\Omega$}
  \label{tab:equations_cgs}
  \centering
  \begin{tabular}{ll}
    \toprule
    Model    &  Unnormalized posterior density $g(\Omega)$    \\
    \midrule
    \vspace{0.2cm}
    LDA & $(C^{-j}_{k_j,d_j}+\alpha) \dfrac{D^{-j}_{k_j,w_j}+\beta}{D^{-j}_{k_j,.}+V\beta}$      \\
    \vspace{0.2cm}
    RTM  &$(C^{-j}_{k_j,d_j}+\alpha) \dfrac{D^{-j}_{k_j,w_j}+\beta}{D^{-j}_{k_j,.}+V\beta}exp\left(\dfrac{b_k}{N_d}\sum \dfrac{n_{d',k}}{N_{d'}}\right)$\\
    sLDA & $(C^{-j}_{k_j,d_j}+\alpha) \dfrac{D^{-j}_{k_j,w_j}+\beta}{D^{-j}_{k_j,.}+V\beta}exp\left(-(y_d-b^T\bar{z}_d-a)^2\right)$   \\
    \bottomrule
  \end{tabular}
\end{table}

%\begin{equation}
%\label{eq:posteriorcgs}
%p(Z_j=k|x,z_{-j},\alpha,\beta) \propto (C^{-j}_{k_j,d_j}+\alpha) %\dfrac{D^{-j}_{k_j,w_j}+\beta}{D^{-j}_{k_j,.}+V\beta}
%\end{equation}

The unified CGS algorithm shares some commonality across several graphical models. 
It first computes the statistics $\Omega=\{C,D\}$ for each model, which are required to compute the posterior equation $g(\Omega)$ for each record in the dataset. 
%For illustration, the statistics of LDA correspond to  $\Omega=\{C,D\}$. 
Rather than recomputing all the parameters of the statistics for every record in the dataset, it is more desired and more efficient to just update the statistics as required. 
In step 5, CGS subtracts the contribution of the $j^{th}$ record from each of the statistics 
before computing the conditional distribution $g(\Omega)$. 
Then, at step 8, it computes the statistics with the updated assignment $z'_j$. 

Equation \ref{eq:updatestats} shows the update for the statistics of the LDA model 
at the $j^{th}$ record, corresponding to the function \textit{updateStats} in Algorithm \ref{alg:cgs}:

\begin{equation}
\label{eq:updatestats}
C_{k_jd_j} = C_{k_jd_j} + 1 ~~;~~ D_{k_jw_j} = D_{k_jw_j} + 1
\end{equation}
%\begin{equation}
%\label{eq:updatestats}
%C_{z'_jd_j} = C_{z'_jd_j} + 1 ~~;~~ D_{z'_jw_j} = D_{z'_jw_j} + 1
%\end{equation}

%\vspace{1cm}

This algorithm explicitly reveals problems caused by Gibbs sampling. 
First, the statistics used to compute the posterior in Line 6 are updated after every tuple in the dataset. As a consequence, the algorithm is non-distributable.
%non-trivially distributable. 
Though approximate methods have been proposed to solve the distributability of the inference, 
they are only applicable to some specific graphical models, and sacrifice the predictive power of the collapsed Gibbs sampler \cite{paralelllda,GLDA}.
In addition, Line 6 and 7 show that the posterior must be computed and sampled for every row in the dataset, even when the content of multiple observed tuples is repeated. 
For example, the same word dould appear multiple times in a document in reality. 
The repetition of tuples can considerably increase the computational complexity. 
For instance, the Cora dataset contains 23.01\% repeated tuples, while the Diggs repeats 18.17\% of its tuples.

\begin{figure}
  \includegraphics[width=\linewidth]{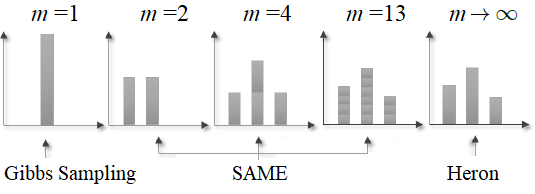}
  \caption{Approximation of the posterior $p(Z_j|Z_{-j}={z_{-j}},\mathbb{D})$) via sampling. 
  The figure shows examples of $n$ draws from the posterior. As the number of samples grows, we obtain better estimate of the distribution. The samples approach the posterior itself when $m$ grows up to infinity. }
  \label{fig:repeated}
\end{figure}

%Cool LDA
To cope with these problems, a state augmentation inference method, 
called state augmentation for marginal estimaton (SAME),
introduces the idea of combining multiple independent samples of each posterior \cite{cooled}. This effectively corresponds to ``cooling'' the posterior, allowing an annealed search of the MAP parameters. As a result, the inference often yields higher-quality estimates than the single sample based approaches. 
To deal with the higher cost of obtaining additional samples, 
SAME employs a coordinated-factored approximation, in which it replaces sampling $m$ multinomial distributions with a single Poisson sample per category, hence the label of coordinated approximation. 
The efficiency of a Poisson sampler remains similar to sampling from a multinomial distribution as long as $m$ takes a value of $100$ or less. 
The sampling scheme replaces Line 7 and 8 in Algorithm \ref{alg:cgs}, as shown below:
\begin{flalign}
\label{eq:cooled}
\nonumber
 z'_j(k) \sim Poisson(~m*p(Z_j\!=\!k|Z_{-j}\!=\!{z_{-j}},\mathbb{D})~) ~\forall k \in 1 \cdots K \\ 
\dot{C}_{k,d_j}=\dot{C}_{k,d_j}+z'_j(k)/m ~~;~~ \dot{D}_{k,w_j}=\dot{D}_{k,w_j}+z'_j(k)/m
\end{flalign}

%% DAN: This may be moved to next section @ Feb 12, 2:28pm
%% Under construction
We note that the statistics of the method take into account the $n^{th}$ replication of the latent state as follows:

\begin{equation}
\dot{C}_{k,d_j}=\dfrac{1}{m}\sum^m_{n=1} C^{(n)}_{k,d_j} ~~;~~  \dot{D}_{k,w_j}=\dfrac{1}{m}\sum^m_{n=1} D^{(n)}_{k,w_j}
\end{equation}

\section{Heron Inference}

Following SAME, the proposed Heron inference method leverages multiple samples to ``cool'' the posterior. Differently, Heron introduces a new mathematical manipulation to drive the number of samples $m$ up to infinity, yielding a deterministic fixed-point algorithm. %The advantages are potential improved accuracy over cooled LDA, improved speed, guaranteed parallelism.  

We follow three main steps to derive Heron inference method. 
Firstly, we show that by maximizing the replication of samples, we obtain an update that is no longer random,
i.e., transforming Gibbs sampling into a deterministic algorithm. 
Secondly, we study the convergence properties of the novel deterministic algorithm, and realize that the method is implicitly solving a fixed-point system of equations, which presents a new framework for doing inference for probabilistic graphical models. %The proposed method exchanges the topic allocation for the allocation of a  convex mixture of topics providing a higher flexibility which results in higher predictability.
Lastly, we observe that repeated document-word tuples converge to the same fixed-point equations, hence they only need to be solved once by the algorithm. %As result the algorithm eliminates the need for computing posteriors of duplicate tuples yielding a decrease on the computational complexity. 

%steps:
%1) deterministic update
%2) system of equations at convergence
%3) repeated tuples converge to the same equations

First, the two steps in Equation \ref{eq:cooled} can be reduced to one step by deriving the distribution from the linear transformation of the Poisson distribution when multiplied by $\dfrac{1}{m}$. Let $Y_k$ be a $Poisson(mp(Z_j=k|Z_{-j}{=}{z_{-j}},\mathbb{D}))$ random variable, and let $U_k$ represent a linear transformation of $Y_k$ such that $U_k=\dfrac{1}{m}Y_k$. 
Then, the probability mass function of $U_k$ can be expressed in terms of any outcome $u$ as :

\begin{equation}
p(U_k) = \dfrac{(mp(Z_j=k|Z_{-j}={z_{-j}},\mathbb{D}) )^{mu}e^{-mp(Z_j=k|Z_{-j}={z_{-j}},\mathbb{D}) }}{mu!}
\end{equation}

As we increase $m$, we observe that the variance decreases. In effect the variance tends to zero as $m$ grows up to infinity. Instead, the mean for each outcome $u$ remains as:

\begin{equation}
\begin{aligned}
%Var(U)=\dfrac{1}{m}p(Z_j|Z_{-j}={z_{-j}}) \\
Mean(U_k)=p(Z_j=k|Z_{-j}={z_{-j}}) = \lambda 
\end{aligned}
\end{equation}

\begin{figure}[h]
\includegraphics[width=8cm]{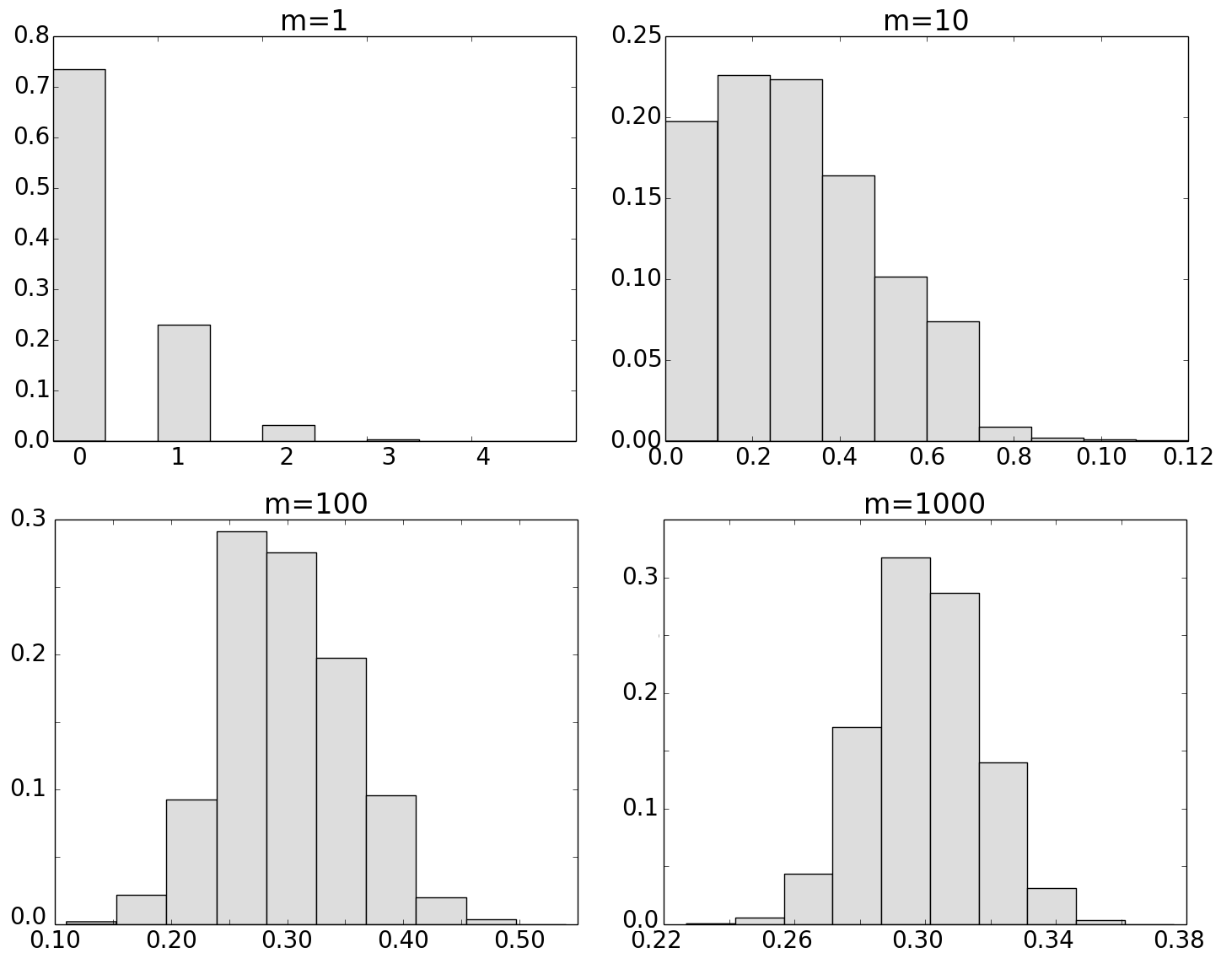}
\caption{Behavior of $p(U_k)$ as  $m\to\infty$ for $\lambda =0.3$}
\label{fig:minfty}
\end{figure}

The probability mass starts to concentrate in the mean value  $\lambda$. Figure \ref{fig:minfty} illustrates this effect.

%Analytically, for a given $\epsilon > 0$, there exists $y$ such as the distribution in $y$ has mean $\lambda$ and variance < $\epsilon$
%To complete the first step of Heron method, we let $m$ tend to infinity and we observe that the distribution p
%$$lim m infty  \dfrac{(m\lambda)^{my}e^{-m\lambda}}{my!} = 1 \lambda if y=\lambda$$

As a result, by maximizing $m$, we obtain an update that does not require any sampling.
Then, Line 7 and 8 of Algorithm \ref{alg:cgs} can be replaced by: 
\begin{equation}
\label{eq:dgmupdate}
\begin{aligned}
\ddot{C}_{k,d_j}= \ddot{C}_{k,d_j} +p(Z_j=k|Z_{-j}={z_{-j}},\mathbb{D}) \\
\ddot{D}_{k,w_j}= \ddot{D}_{k,w_j} + p(Z_j=k|Z_{-j}={z_{-j}},\mathbb{D})
\end{aligned}
\end{equation}
where 
\begin{equation}
\label{eq:dgmstats}
\begin{aligned}
\ddot{C}_{k,d_j}=\sum_{x \in d_j}p(Z_j=k|Z_{-j}={z_{-j}},\mathbb{D}) \\
\ddot{D}_{k,w_j}=\sum_{x \in w_j} p(Z_j=k|Z_{-j}={z_{-j}},\mathbb{D}) 
\end{aligned}
\end{equation}

%m independent distributions  The update of the sampler corresponds to a coordinate factored Dirichlet-Multinomial Compound distribution \cite{cooled}. Equation \ref{eq:cooled} shows the update for LDA. Similar derivations can be obtained for RTM and SLDA. 

%\begin{equation}
%\label{eq:cooled}
%p_{Z_j}= \left(\sum C^{-j}_{k_j,d_j} / m+\alpha\right) \dfrac{ \sum D^{-j}_{k_j,w_j} / m+\beta}{\sum D^{-j}_{k_j,.} / m+V\beta}
%\end{equation}

%By taking the limit to infinity we obtain the following update:

%$lim_{m \to \infty} p_{Z_j}=  \left(C^{-j}_{k_j,d_j} +\alpha\right) \dfrac{D^{-j}_{k_j,w_j} +\beta}{D^{-j}_{k_j,.}+V\beta}$ \\

%Now that we have a deterministic algorithm,
Second, we examine the convergence properties of Gibbs sampling. As shown by \cite{casella}, Gibbs sampling converges to the stationary distribution of a Markov chain, where the transition matrix corresponds to the multiplication of the conditional distributions  $\prod^{|X|}_{m=1} p(Z_m|Z_{-m},X)$. Hence, labeling $P(Z^{*}_j)$ as the $j^{th}$ stationary distribution, we have:

\begin{equation}
\label{eq:system}
p(Z^{*}_j)=p(Z^{*}_j) \prod^{|\mathbb{D}|}_{m=1} p(Z_m|Z^*_{-m}=z^*_{-j},\mathbb{D}) ~\forall j \in 1 \cdots |\mathbb{D}|
\end{equation}

Estimation of these marginal distributions is based merely on the topic assignments $z_j$, $~\forall j \in 1 \cdots |\mathbb{D}|$. Hence, upon convergence, there must exist a set of values $z^*_j$ that satisfy or approximate the system of Equations \ref{eq:system}. %Because of the random effects of sampling, it is not possible converge to a fixed set of values $z_j$. 
Furthermore, the solution to the system of equations presented in Equation \ref{eq:system} as necessary condition approximates the conditional distributions $p(Z^{*}_j|Z^{*}_{-j}={z^{*}_{-j}},\mathbb{D})$ to the marginal distributions $p(Z^{*}_j)$. Therefore, we obtain:

\begin{equation}
\label{eq:semifix}
p(Z^*_j) \approx p(Z_j|Z^{*}_{-j}=z_{-j},\mathbb{D})=g(z_{-j})
\end{equation}
%\subsection{2. By splitting each record infinitesimally we obtain an explicit fixed-point formulation}
where $g(\Omega)$ is the posterior equation derived for the model of interest (see Table \ref{tab:equations_cgs}). Note that $\Omega$ is computed based on the topic assignments $z$. To transform Equation \ref{eq:semifix} into a fixed-point equation, we need to find a representation such that the posterior is written in terms of deterministic updates %$p(Z^*_{-j})=g(p(Z^*_{-j}))$. 
as opposed of the random topic assignments of the collapsed Gibbs sampler.

%%This equation shoukd be DGM one.
%% P(Z|Z=z,X)=(C+\alpha) (D+\beta)

To that end, we use the deterministic Equation \ref{eq:dgmupdate} obtained above. This equation explicitly expresses how the update of $p(Z_j)$ is given by the parameters of the posterior equation. 
As a result, we obtain the fixed-point condition that must hold at convergence of the given model. 
To simplify the notation, we define $x_{dw} \equiv p(Z_j=k|Z_{-j}={z_{-j}},\mathbb{D})$, 
where $j$ indexes the document-word tuple composed by $(d,w)$.

\begin{equation}
\label{eq:fixpoint}
x_{ij}=\dfrac{\dfrac{\left(D_i-x_{ij}+\alpha\right)\left(W_j-x_{ij}\right) }{Z-x_{ij}}}{\sum_k^{|K|}\dfrac{\left(D_{ik}-x_{ijk}+\alpha\right)\left(W_{jk}-x_{ijk}+\beta\right) }{Z_k-x_{ijk}}}\\
\end{equation}
where
\[
D_i\!=\!\sum^{|W|}_j x_{ij}k_{ij} ~~;~~ W_j\!=\!\sum^{|D|}_i x_{ij}k_{ij}
 ~~;~~ Z\!=\!\sum^{|W|}_j\sum^{|D|}_i x_{ij}k_{ij}
\]

%For that, we propose we propose splitting each record to obtain an approximation of the posterior distributions $p(Z^{*}_j|Z^{*}_{-j}={z^{*}_{-j}},X)$. While in Gibbs sampling a single sample is used to condition on the posteriors, we find that sampling n times improves the information used on the conditional. For example, in Figure \ref{fig:sample}, we show how by sampling n times we can approximate the posterior distribution. In particular, by sampling an infinite number of times. The law of the large numbers gives us a guaranteed estimation of the distribution itself (Equation \ref{eq:lln})

%\begin{equation}
%\label{eq:lln}
%equation LLN
%\end{equation}

%In this case, the posterior equation of LDA presented in Equation 1, transforma into:

%\begin{equation}
%\label{eq:dgm}
%equation DGM
%\end{equation}

%This equation 

%\[
%x_{ij}=\dfrac{\dfrac{\left(D_i-x_{ij}\right)\left(W_j-x_{ij}\right) }{Z-x_{ij}}}{\sum_k^{|K|}\dfrac{\left(D_{ik}-x_{ijk}\right)\left(W_{jk}-x_{ijk}\right) }{Z_k-x_{ijk}}}
%\]

%In the inference problem, it is sufficient to finding $P_z$ at convergence. That is because, we $P_z$ can be used to estimate $\Theta$ and $\Phi$ by using:

%--
We have transformed the inference problem into a system  of $|D| \times |Nd|$  non-linear equations whose solutions correspond to all possible outcomes of an infinitely replicated latent state augmentation independent of the initialization. 
%Equation \ref{eq:dgm} corresponds to the update of each row i. In the Gibbs Sampling formulation, each posterior correspond to a component of the trans this is equivalent to the 

%Equation 6 corresponds to the update of each row i. As the number of iterations approach to infinity, the posterior distributions will converge and so we will have that:

%\[
%D_i=\sum^{|W|}_j x_{ij}k_{ij}
%\]

%\[
%W_j=\sum^{|D|}_i x_{ij}k_{ij}
%\]

%\[
%Z= \sum^{|W|}_j\sum^{|D|}_i x_{ij}k_{ij}
%\]

%\subsection{3. The posterior of repeated tuples converges to the same distribution}

Finally, we can further simplify this system of equations by observing that the same document-word tuples will also converge to the same values. Thus, we have:

\begin{equation}
\begin{aligned}
\label{eq:repeatedtuples}
P(Z_j|Z_{-j},X)=P(Z_m|Z_{-m},\mathbb{D}) ~\forall j \in 1 ..|\mathbb{D}|\\ \text{such that } d_j=d_m \wedge w_j=w_m
\end{aligned}
\end{equation}

This is easy to check using Equation \ref{eq:fixpoint}. Upon convergence, repeated tuples will have the same fixed-point equations. Now the problem has been transformed into solving a system of $|D|\times|W|$ non-linear equations. As a solution, we leverage the fixed-point iterative method. The fixed-point method finds the solution to a system of equations $x=f(x)$ as long as $f$ is a continuous function. 

The algorithm is similar to CGS presented in Algorithm \ref{alg:cgs} with the main differences:
1) There is no need to iterate over $|D|\times|Nd|$ steps, and 2) The algorithm is distributable since we only need to update the statistics until the end of each iteration. 
Theoretically, we can partition the dataset into as many pieces as $|D|\times |W|$. The end result will be the same as long as we combine the results at the end of each iteration. We present Heron inference in Algorithm \ref{alg:fixed}. Note that the definition of $\Omega=\{C,D\}$ in \textit{computeStats} is overridden with $\ddot{\Omega}=\{\ddot{C},\ddot{D}\}$.

\begin{algorithm}
   \caption{Heron Inference }
   \label{alg:fixed}
\begin{algorithmic}[1]
   \STATE {\bfseries Input:} data $x_j$, latent assignment $z_j$, dataset size $m$, number of repeated tuples $\rho_j$
   \STATE \textbf{Initialize} $z_j$ 
   \STATE $\ddot{\Omega}$= computeStats($z_j$,$\rho_j$) from Equation \ref{eq:dgmstats}
   \REPEAT
   \FORALLP{$d=1$ {\bfseries to} $|D|$ }
     \FORALLP{$w=1$ {\bfseries to} $|W|$ }
  		 \STATE $x_{dw} \propto g(\ddot{\Omega})$ from Table \ref{tab:equations_cgs}
    \ENDFAP
   \ENDFAP
   \STATE $stats$= computeStats($z_j$,$\rho_j$) from Equation \ref{eq:dgmstats}
   \UNTIL{convergence} (Equation \ref{eq:KL})
\end{algorithmic}
\end{algorithm}

An additional advantage of the deterministic Heron method is that the assessment for convergence is now possible within the Gibbs framework. We are allowed to explore the assessment for convergence using the average Kullback-Leibler divergence (KL) or the average Chi-Squared distance on the learned $\theta$ and $\phi$. We find that the latter metric oscillates and it does not have a monotonic behavior. We find that it is often more practical to measure the change in either $\theta$ or $\phi$. Equation \ref{eq:KL} presents the metric used to assess convergence of Heron at the $i^{th}$ iteration.

\begin{equation}
\label{eq:KL}
D_{KL}(\theta^i||\theta^{i-1})= \dfrac{1}{|\mathbb{D}|} \sum^{|\mathbb{D}|}_{d=1} \theta^i_d ~ log(\dfrac{\theta^i_d}{\theta^{i-1}_{d2}}) 
\end{equation}
%-- Note on convergence or matrix caching.

Finally, the SAME based methods are under the premise of independence among topics, therefore the methods such as hidden Markov model \cite{HMM} and stochastic block mixture model \cite{SBMM} are not guaranteed to work using these inference approaches.

\section{Experiments}
\label{sec:Experiments}

%%RQs
We evaluate our proposed Heron inference method against two well-established baseline methods. In this section we aim to answer the following research questions (\textbf{RQs})

\begin{description}
\item[RQ1] Is Heron able to fit better a test set than the state-of-the-art inference method SAME?
\item[RQ2] What is the effect on perplexity as we increase the number of batches, and does the moving average estimate of $\theta$ improve Heron's learning?
\item[RQ3] How sensitive is Heron to the setting of the hyperparameters $\alpha$ and $\beta$?
\item[RQ4] How does the efficiency of Heron compare against the arguably fastest GPU-based inference method within the Gibbs framework?
\item[RQ5] How does the efficiency of Heron inference get affected by the batch size?
\item[RQ6] Does Heron help extract topics with acceptable coherence?
\end{description}

%% Datasets
\textbf{Datasets} We used three popular and publicly available
datasets for our experimental evaluation. 

\begin{itemize}
%\item \textbf{NIPS\footnote{http://ai.stanford.edu/~gal/data.html}} \cite{chechik2007eec} is a dataset that contains the co-occurrence of words and authors from the NIPS\footnote{\url{https://nips.cc/}} (Neural Information Processing Systems) proceedings from 1988 to 2003.  NIPS (Large) with 1265 authors and 2245 words producing 4218493 records and NIPS (Small) with 731 authors and 1853 words producing 3508065 records. 

%\item \textbf{LastFM\footnote{\url{https://www.last.fm/}}} is a popular music streaming service. We use the LastFM (360K) dataset\footnote{\url{http://www.dtic.upf.edu/~ocelma/MusicRecommendationDataset/lastfm-360K.html}} containing (user, artist, plays) tuples for 360K users. We also split this dataset into two different sizes. Lastfm (Large) with 22567 users and 5185 artists producing 145534518 records and Lastfm (Small) with 3079 users and 2883 artists producing 12268186 records. 

\item \textbf{Cora} dateset contains research papers and their citation networks.
%used to enforce relations among papers. 
After preprocessing, we obtained $2,618$ documents with a vocabulary composed of 1,400 words. 
The number of citation links is 5,212. 
The dataset was first used for testing the RTM model \cite{RTM} \footnote{\url{https://people.cs.umass.edu/~mccallum/data.html}}.
\item \textbf{Diggs} dataset
%is a popular community, where users shares the links to webpages with short descriptions, and other users in %the community can ``digg'' the links that they like. 
%and we use the number of ``diggs'' received per link as the link's rating. 
consists of $6,192$ document-rating pairs with a vocabulary composed of 3,145 words after preprocessing.
Note that, 
following \cite{sLDA}, 
%given the regression function by Equation \ref{eq:super}, it is convenient to 
we normalized the ratings into the range $[1,5]$ to avoid numerical overflows when calculating sLDA's posterior. 
The dataset was originally used for evaluating sLDA \cite{sLDA} 
\footnote{\url{http://www.cs.columbia.edu/~blei/papers/digg-data.tgz}}.
\item \textbf{Movielens} is a popular benchmark for rating prediction and recommendation. 
We used the MovieLens 20M dataset \footnote{\url{https://grouplens.org/datasets/movielens/}},
and focused on the most popular 100,000 users who rated 25,646 movies. 
We treated the movies and users as pseudo documents and words, respectively. 
The average rating of all the related users on a movie was used as the rating of the movie (document).
\end{itemize}

% ~~ Hyperparameters ~~ %

% ~~ Baselines ~~ %
\textbf{Baselines} 
%In order to ascertain the effectiveness of our proposed approach, 
We compared Heron inference with two strong baselines, i.e., standard collapsed Gibbs sampling and SAME inference method \cite{cooled}.
Collapsed Gibbs sampling is perhaps one of the most commonly used methods for inferrring graphical models, 
while SAME has been shown to attain the best predictive perplexity for inferring LDA model.
%possibly due to the state augmentation based inference and the fastest learning time given to its Poisson %based sampling inference. 

% ~~ Models ~~ %
\textbf{Topic Models} 
To test the performance of Heron for inferring graphical models, we used three well-known topic models, i.e., LDA (Latent Dirichlet Allocation) \cite{LDA}, RTM (Relational Topic Model) \cite{RTM} and sLDA (Supervised Latent Dirichlet Allocation) \cite{sLDA}. 
%Among these, SAME-LDA implementation is open source \footnote{\url{https://github.com/BIDData/BIDMach}}. We %derived and implemented SAME-RTM and SAME-sLDA. 
For LDA, all the datasets can be used to evaluate the inference methods. 
Inference for RTM can be only tested on the Cora dataset, 
as only Cora has a citation networks.
We used Movielens and Diggs to test the inference for sLDA,
since in addition to documents/movies, corresponding ratings are also available.
%has an associated ranking.  

\textbf{Implementation Details} All the algorithms were evaluated on a single PC equipped with a single 2-core CPU (Intel Xeon @ 2.20 GHz) and a \textit{Nvidia tesla K80} GPU with dual stream processors. Each GPU processor comes with $2,496$ cores and $12$GB of GDDR5 clocked at $5$GHz. 
Only one processor of the GPU was used in the benchmark. 
The standard Gibbs sampler was implemented using CPU resources. 
In all our experiments we fixed both the grid size and block size.

% ~~ Experiments ~~ %
 We assess our proposed inference method in terms of running time, predictability, as well as coherence of the extracted topics. In particular, we conduct the following three experiments. 
 
 \begin{table}[h]
\caption{Hyperparameter setting per dataset.}
\label{tab:hyperparameters}
\centering
\begin{tabular}{lcccc}
\toprule
Dataset      & $\alpha$ & $\beta$ & $\eta$ & $a$\\ 
\midrule
Cora         & 0.4 & 0.3 & 0.75 &  N/A  \\
Diggs        & 0.6 & 0.5 & 0.5 &  0.25  \\
Movielens  & 0.6 & 0.6 & 0.5 &  0.4 \\
\bottomrule
\end{tabular}
\end{table}

\begin{table}
\caption{Predictive perplexity versus the nunber of topics.}
\label{tab:perpK}
\centering
\begin{tabular}{ll@l^r^r^r^r}
\toprule
 & \multirow{ 2}{*}{Model} & \multirow{ 2}{*}{Inference} &\multicolumn{4}{c}{Perplexity}\\
				  &       &           & K=10 & K=25 & K=50 & K=100 \\
\midrule
\multirow{ 6}{*}{\rotatebox{90}{Cora}}      & \multirow{ 3}{*}{LDA}    & \rowstyle{\cellcolor{hcolor1}}CGS  & 1709.63 & 1837.46 & 1999.25 & 2623.86 \\
                            &       &  \rowstyle{\cellcolor{hcolor2}}SAME                 & 1706.27 & 1821.20 & 1977.26 & 2549.43  \\
                            &       & \rowstyle{\cellcolor{hcolor3}}Heron                   & 1683.03 & 1770.66 & 1977.51 & 2550.21 \\
                            & \multirow{ 3}{*}{RTM}    & \rowstyle{\cellcolor{hcolor1}}CGS  & 1700.12 & 1830.02 & 1995.37 & 2245.96 \\
                            &       & \rowstyle{\cellcolor{hcolor2}}SAME                  & 1699.20 & 1811.96 & 1991.12 & 2043.35    \\
                            &       & \rowstyle{\cellcolor{hcolor3}}Heron                   & 1682.32 & 1806.09 & 1991.02 & 1971.07  \\
                            \midrule
\multirow{ 6}{*}{\rotatebox{90}{Diggs}}   &   \multirow{ 3}{*}{LDA}   &\rowstyle{\cellcolor{hcolor1}} CGS  & 952.18  & 1166.93 & 1500.51 & 2075.65 \\ 
                            &       & \rowstyle{\cellcolor{hcolor2}}SAME                 & 948.57  & 1161.40 & 1492.53 & 2067.24 \\        
                            &       & \rowstyle{\cellcolor{hcolor3}}Heron                  & 947.27  & 1160.35 & 1489.43 & 2062.12  \\   
                            & \multirow{ 3}{*}{sLDA}   & \rowstyle{\cellcolor{hcolor1}}CGS & 940.09  & 1125.85 & 1452.94 & 2007.00 \\ 
                            &       & \rowstyle{\cellcolor{hcolor2}}SAME                 & 919.31  & 1123.82 & 1446.41 & 2005.28\\  
                            &       & \rowstyle{\cellcolor{hcolor3}}Heron                  & 914.77  & 1121.96 & 1442.54 & 2002.61  \\       
                            \midrule
\multirow{ 6}{*}{\rotatebox{90}{Movielens}} & \multirow{ 3}{*}{LDA}   & \rowstyle{\cellcolor{hcolor1}}CGS  & 1085.47 & 1022.25 &  1011.62 &  1065.02 \\
                            &       & \rowstyle{\cellcolor{hcolor2}}SAME                 &  924.12 &  932.43 &  993.81  &  1045.75   \\
                            &       & \rowstyle{\cellcolor{hcolor3}}Heron                  &  922.20 &  931.34 &  989.92  &  1045.09  \\
                            & \multirow{ 3}{*}{sLDA}  & \rowstyle{\cellcolor{hcolor1}}CGS  &  926.55 &  905.56 &  924.75  &   963.45 \\
                            &       & \rowstyle{\cellcolor{hcolor2}}SAME                 &  923.28 &  903.63 &  921.26  &   963.65  \\
                            &       & \rowstyle{\cellcolor{hcolor3}}Heron                  &  920.22 &  899.68 &  918.51  &   954.28  \\
\bottomrule
\end{tabular}
\end{table}

\subsection{Predictive Perplexity}

% ~~ Problem Definition and settings~~ %
The problem defined is given as follows: given a fraction of the original data, we evaluate our model's ability to generate the withheld portion. As such, the evaluation metric adopted in our experiments is the predictive perplexity. The number of topics $K$ is tuned amongst $\{10,25,50,100\}$. The training-testing split is $70-30$, and we run all models for $1,000$ iterations. Furthermore, we conducted five trials for each setting in the spirit of robust experimentation, and we reported the average results. The hyperparameters are given in Table \ref{tab:hyperparameters}. Models applied to the same dataset use the same hyperparameter settings.

\begin{figure}
  
 \subfloat{\includegraphics[width=0.24\textwidth]{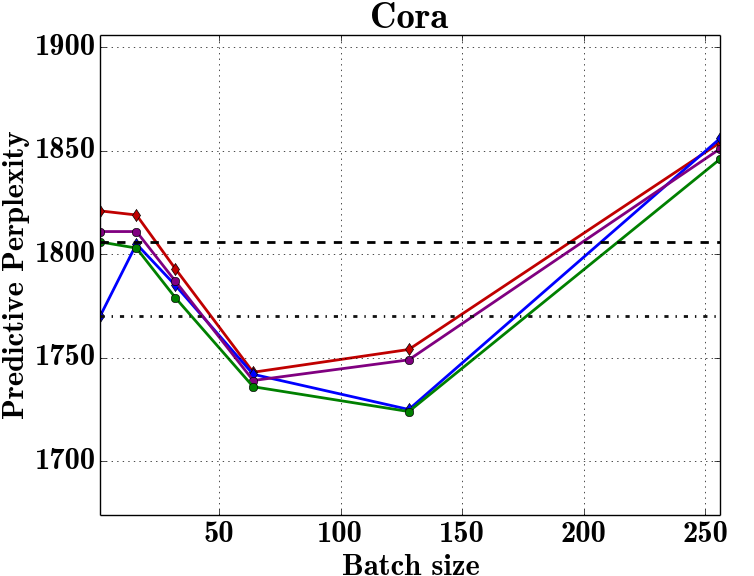}\label{fig:lastfmlarge}}
~~~~\subfloat{\includegraphics[width=0.24\textwidth]{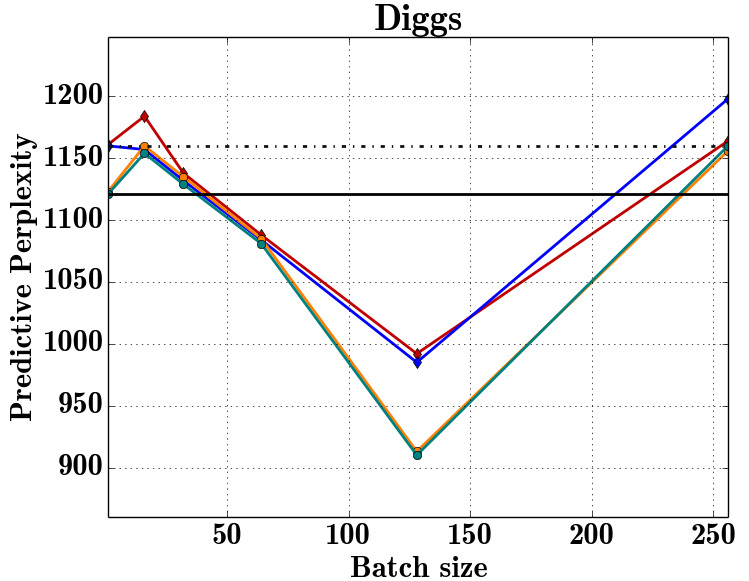}\label{fig:movielens}}  
 
\subfloat{\includegraphics[width=0.24\textwidth]{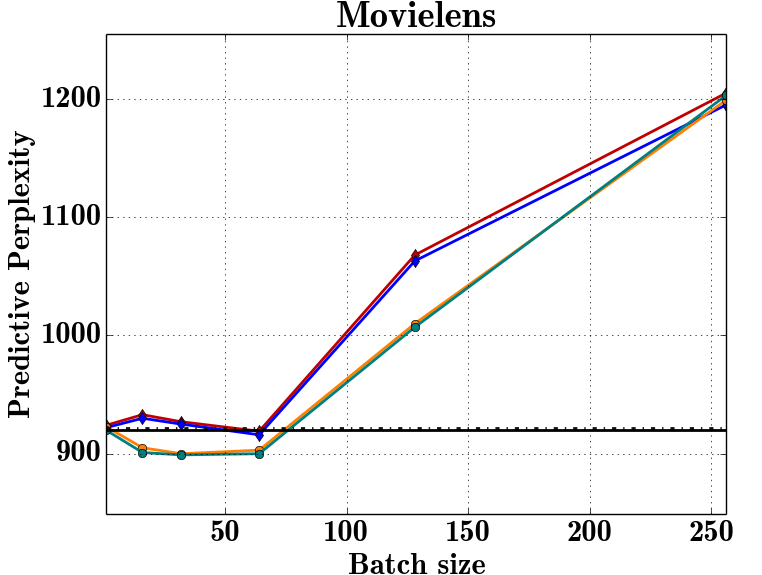}\label{fig:nipslarge}}
~~~~~~\subfloat{\includegraphics[width=0.24\textwidth]{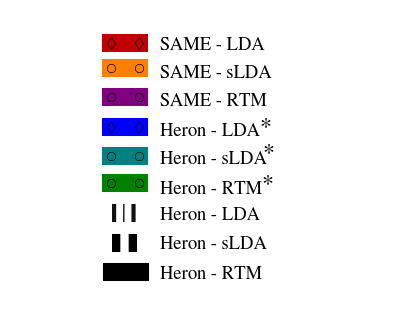}}

  \caption{Predictive perplexity on the GPU based inference methods versus the number of batches.}
  \label{fig:batches}
\end{figure}

Table \ref{tab:perpK} shows the comparison results of all the inference methods when applied to the three models. Answering \textbf{RQ1}, our proposed Heron inference outperforms CGS and SAME method, i.e., both the best and second best performance on each configuration comes from Heron model across all $K$ values. %Additionally, when comparing both techniques with same $K$ values, Heron has a better predictive power than SAME method. 
As expected, SAME method consistently outperforms CGS. This is due to the fact that SAME has been derived from CGS by applying a finite replication of the latent states, and Heron maximizes this effect using a mathematical manipulation. SAME method improves Gibbs sampling because the estimation for the parameters is done based on the document-topic and word-topic counts. Hence, the estimation is based on discrete values. By using multiple replications of the latent state, it is possible to make this estimation more granular so that the counts can be thought to reside in the space of real numbers. As we increase the number of replications, the estimation becomes more granular, and Heron becomes the most flexible approach approximating the local optima closer than the other methods.

%In addition, we observe that the ratio between the number of document-word tuples and the dimensionality of the document-word matrix reveals an estimation of the gain in perplexity when switching from CGS to Heron inference.  Datasets with lower ratio values will benefit from a better improvement when choosing Heron method over CGS. For instance, we observe a higher improvement in Movielens having a ratio of xx samples per dimension. Instead, the improvement is relatively smaller in Cora and Diggs with ratios of xx and xx samples per dimension respectively. The rationale is that datasets with lower samples per dimension will have less samples per document and less samples per word in the vocabulary so that CGS will not achieve a good estimation of the model's parameters.

The benefit of the distributed methods lies in their ability to split the dataset into batches, so that several GPUs can be used to boost the speed of inference. As such, we propose \textbf{RQ2} and solve it by studying the effect of splitting the datasets into different batch sizes. We fix $K=25$ and split the dataset into exponentially larger number of batches. Similar to SAME method, Heron inference may treat the batches as an infinite stream, and it is not necessary to denote passes over the dataset. In addition, we also study SAME's moving average estimate for $\theta$ \cite{cooled}, which has shown improvements when applied to other learning algorithms \cite{ovlda}. We label the results of applying this technique as Heron-LDA$^*$, Heron-sLDA$^*$, and Heron-RTM$^*$.
%The total number of streams required to reach convergence are nearly the same when comparing SAME and Heron Inference.

%\input{tables/perpBatch}

We answer \textbf{RQ2} using Figure \ref{fig:batches}. We conclude that the batched versions converge to a better estimation as measured by the predictive perplexity. As the dataset is split into smaller pieces, the predictive perplexity improves. This may be due to the fact that splitting the dataset into batches allows the inference to reach a better local optima with the use of a moving average estimate for $\theta$ \cite{cooled,ovlda}. Nonetheless, after a certain number of batches the predictive perplexity is affected, indicating that the learning has become compromised. Note that Heron can be distributed into as many batches as the number of document-word tuples. The final inference will result in the same estimation as if we partitioned the dataset into any number of pieces or if we did not partition the dataset. This is because Heron inference is ultimately solving a fixed-point system of equations, given by Equation \ref{eq:fixpoint}. The results show that Heron can yield better local optima if we decide to use the moving average updates. In that case, the learning of Heron may also be compromised if we split the dataset into a large number of batches.

\begin{table}[h]
\caption{Predictive perplexity versus the hyperparameters $\alpha$ and $\beta$.}
\label{tab:sensitivity}
\centering
\begin{tabular}{ll@c^c^c^c^c}
\toprule
\multicolumn{2}{l}{\textsc{LDA - Cora}} &  $\beta=0.1$ & $\beta=0.2$ & $\beta=0.3$ & $\beta=0.4$ & $\beta=0.5$\\
\midrule
\multirow{5}*{\rotatebox{90}{CGS}}&$\alpha=0.1$ & \rowstyle{\cellcolor{hcolor1}}  2118.8 &  1974.42 & 1766.27 & 1741.98 & 1667.46 \\
& $\alpha=0.2$ &\rowstyle{\cellcolor{hcolor1}} 1954.09 & 1831.44 & 1751.84 & 1725.55 & 1677.72 \\
& $\alpha=0.3$ &\rowstyle{\cellcolor{hcolor1}} 1880.51 & 1806.89 & 1747.28 & 1735.68 & 1748.78 \\
& $\alpha=0.4$ &\rowstyle{\cellcolor{hcolor1}} 1912.38 & 1851.29 & 1784.29 & 1784.60 &  1784.55 \\
& $\alpha=0.5$ &\rowstyle{\cellcolor{hcolor1}} 1916.95 & 1882.72 & 1845.47 & 1834.62 & 1845.15 \\
\midrule
\multirow{5}*{\rotatebox{90}{SAME}}&$\alpha=0.1$ &\rowstyle{\cellcolor{hcolor2}} 2102.48 & 1916.14 & 1721.83 & 1649.48 & 1656.24 \\
& $\alpha=0.2$ &\rowstyle{\cellcolor{hcolor2}}1942.4 &  1775.68 & 1713.5 &  1664.11 & 1656.55 \\
& $\alpha=0.3$ &\rowstyle{\cellcolor{hcolor2}}1824.89 & 1794.71 & 1735.07 & 1707.69 & 1719.56 \\
& $\alpha=0.4$ &\rowstyle{\cellcolor{hcolor2}}1840.68 & 1779.86 & 1796.25 & 1787.89 & 1798.68 \\
& $\alpha=0.5$ &\rowstyle{\cellcolor{hcolor2}}1898.44 & 1850.68 & 1845.86 & 1860.4 &  1866.48 \\
\midrule
\multirow{5}*{\rotatebox{90}{Heron}} & $\alpha=0.1$ &\rowstyle{\cellcolor{hcolor3}}1858.19 & 1738.36 & 1666.94 & 1647.12 & 1635.79 \\
& $\alpha=0.2$&\rowstyle{\cellcolor{hcolor3}} 1751.62 & 1704.31 & 1656.84 & 1661.41 & 1656.67 \\
& $\alpha=0.3$&\rowstyle{\cellcolor{hcolor3}} 1708.78 & 1655.16 & 1671.78 & 1692.64 & 1702.59 \\
& $\alpha=0.4$&\rowstyle{\cellcolor{hcolor3}} 1745.76 & 1739.17 & 1737.66 & 1750.46 & 1775.52 \\
& $\alpha=0.5$&\rowstyle{\cellcolor{hcolor3}}1801.18 & 1767.2 &  1782.23 & 1831.43 & 1847.63 \\
\bottomrule
\end{tabular}
\end{table}

As proposed in \textbf{RQ3}, we study the hyperparameter sensitivity of the different inference methods. We made a grid search on $\alpha$ and $\beta$ starting at 0.1 and stopping at 0.5 with steps of 0.1. We applied the inference methods to LDA on the Cora dataset.

Table \ref{tab:sensitivity} shows that the predictive perplexity of Heron is superior to that of SAME, and as expected, SAME has more predictive power than CGS. The results are solid, Heron lower bounds SAME, and SAME lower bounds CGS. Based on the range of the perplexity across different hyperparameters, we conclude that Heron is more robust to the hyperparameter settings, while CGS and SAME have similar sensitivity.

\begin{table}[b]
\caption{Average running time per iteration versus the number of topics.}
\label{tab:timeK}
\centering
\begin{tabular}{ll@l^r^r^r^r}
\toprule
 & \multirow{ 2}{*}{Model} & \multirow{ 2}{*}{Inference} &\multicolumn{4}{c}{Running time ($\mu s$)}\\
				  &       &          & K=10 & K=25 & K=50 & K=100\\
\midrule
\multirow{ 6}{*}{\rotatebox{90}{Cora}}      & \multirow{ 3}{*}{LDA}    & \rowstyle{\cellcolor{hcolor1}}CGS  & 1350 & 1390 & 1460 & 1728 \\
                            &       &  \rowstyle{\cellcolor{hcolor2}}SAME                 &  190 &  214 &  231 & 291 \\
                            &       & \rowstyle{\cellcolor{hcolor3}}Heron                    &    3 &    7 &   11 & 18 \\
                            & \multirow{ 3}{*}{RTM}    & \rowstyle{\cellcolor{hcolor1}}CGS   & 1540 & 1620 & 1690 & 1720 \\
                            &       & \rowstyle{\cellcolor{hcolor2}}SAME                  &  210 &  218 &  222 & 240 \\
                            &       & \rowstyle{\cellcolor{hcolor3}}Heron                  &    5 &   12 &   19 & 33 \\
                            \midrule
\multirow{ 6}{*}{\rotatebox{90}{Diggs}}   &   \multirow{ 3}{*}{LDA}   &\rowstyle{\cellcolor{hcolor1}} CGS  & 2590 & 3140 & 3030  & 3300 \\ 
                            &       & \rowstyle{\cellcolor{hcolor2}}SAME                  &  260 & 290 &  270 & 330 \\        
                            &       & \rowstyle{\cellcolor{hcolor3}}Heron                  &   13 &    15 &   18  &   31 \\   
                            & \multirow{ 3}{*}{sLDA}   & \rowstyle{\cellcolor{hcolor1}}CGS & 4060 & 4090 & 4470 & 5120 \\ 
                            &       & \rowstyle{\cellcolor{hcolor2}}SAME                 &  270 & 280 &  280 &   320 \\  
                            &       & \rowstyle{\cellcolor{hcolor3}}Heron                  &   15 & 23 &   30 &    50 \\       
                            \midrule
\multirow{ 6}{*}{\rotatebox{90}{Movielens}} & \multirow{ 3}{*}{LDA}   & \rowstyle{\cellcolor{hcolor1}}CGS  & 16405 &17356 & 18511 & 21286 \\
                            &       & \rowstyle{\cellcolor{hcolor2}}SAME                &   780 &  790 &   813 &   833   \\
                            &       & \rowstyle{\cellcolor{hcolor3}}Heron                  &    19 &   44 &   62  &   95  \\
                            & \multirow{ 3}{*}{sLDA}  & \rowstyle{\cellcolor{hcolor1}}CGS   &  7053 & 7818 &  8954 & 10367 \\
                            &       & \rowstyle{\cellcolor{hcolor2}}SAME                  &   954 &  976 &   997 &  1028 \\
                            &       & \rowstyle{\cellcolor{hcolor3}}Heron                   &    23 &   46 &    67 &   102 \\
\bottomrule
\end{tabular}
\end{table}

\subsection{Learning Time}

In this section, we report the runtime to convergence of the inference algorithms normalized by the number of iterations.

First, we study the effect of chaning the number of topics on running time, as shown in Table \ref{tab:timeK}. Remarkably, Heron is about one order of magnitude faster than the state-of-the-art SAME method 
across all the values of $K$. 
The main reason lies in Heron does not rely on the sampling operations for inference. 
In contrast, SAME method not only incurs the sampling costs but also is limited at every clock cycle by the available number of samplers within the GPU. 
Benefitting from purely arithmetic operations, Heron does not suffer from the limitations and sampling costs. We also observe that the running time increases linearly as we increase $K$ for the three inference methods. While this increment is larger for Heron, Heron can also benefit from tuning the \textit{grid size} and \textit{block size} (GPU Kernel parameters) to favor different values of $K$. The flexibility of tuning GPU parameters is reduced for SAME, since at its optimal configuration, the system will be still limited by the availability of hardware based samplers. This answers \textbf{RQ4}. %Note that the convergence has a similar rate because the datasets are not split into batches. 

\begin{figure}
 \subfloat{\includegraphics[width=0.24\textwidth]{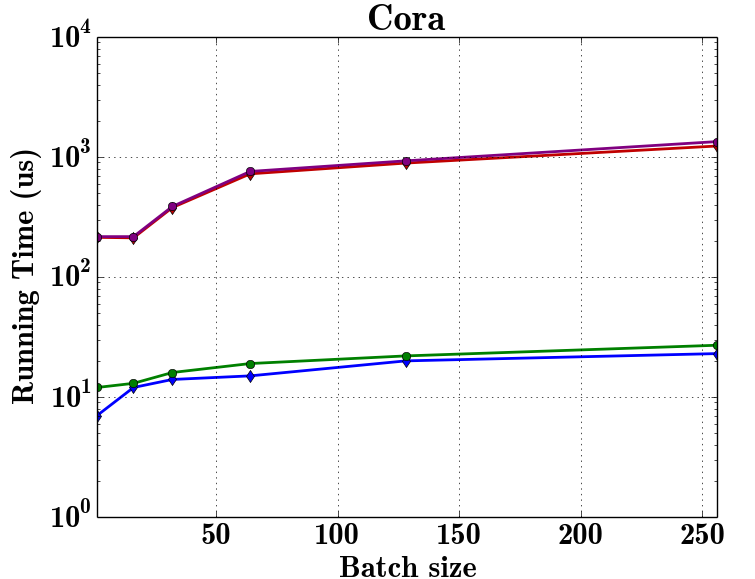}\label{fig:lastfmlarge}}
 ~~~\subfloat{\includegraphics[width=0.24\textwidth]{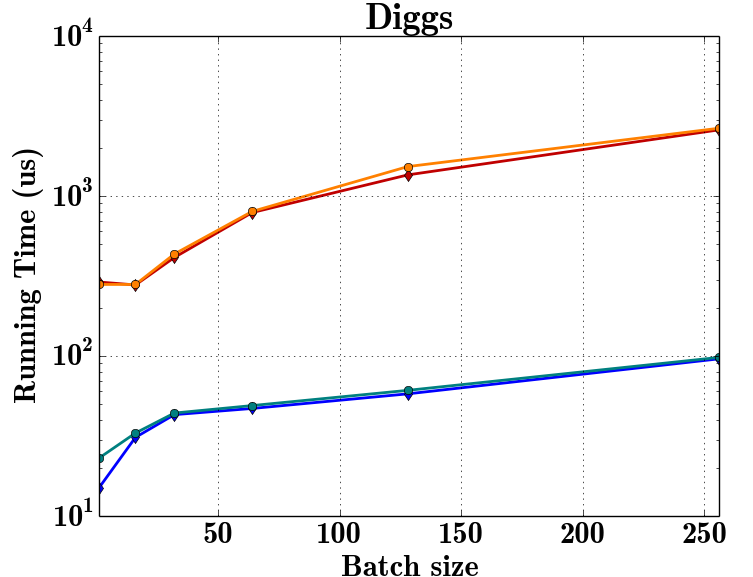}\label{fig:movielens}}  
 
\subfloat{\includegraphics[width=0.24\textwidth]{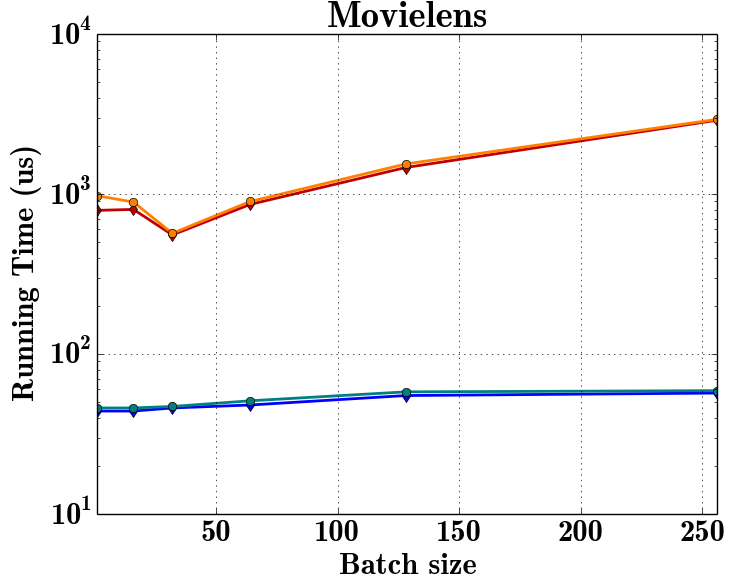}\label{fig:nipslarge}}
~~~~~~\subfloat{\includegraphics[width=0.24\textwidth]{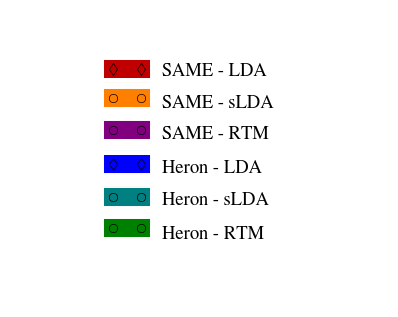}}

  \caption{Average running time per iteration versus the number of batches.}
  \label{fig:timeBatch}
\end{figure}

In Figure \ref{fig:timeBatch}, we show the effect of splitting the datasets into batches. The results show that both inference methods increase its running time logarithmically to the number of batches. This is outstanding given that Heron's convergence is invariant to the number of batches, when the moving average estimate is not applied. It means that Heron can converge to accurate inference results dramatically fast, given that we have enough cores running simultaneously. Answering \textbf{RQ5}, Heron efficiency reduces logarithmically to the number of batches slightly.

%Figure \ref{fig:timeBatch} also shows that the batched version of both methods converges faster than the version without partitioning the dataset. However, the overhead cost of passing data between the GPU and the CPU after each batch involves an overhead cost of around $8~us$ per GPU call for the studied datasets. Hence, it may be preferable to run datasets into smaller number of batches when the number of cores is limited and there is a requirement to optimize for speed.

%\input{tables/timeBatch}

\subsection{Automatic Topic Coherence}

Besides assessing how the methods fit the data, it is important to evaluate the quality of the topics inferred by the methods. Following the evaluation introduced by Newman et al. \cite{autocoherence}, we computed the automatic topic coherence based on the top \textit{N} most likely words for each of the extracted topics. Specifically, we relied on word co-occurrence metrics on the corpus to compute the average coherence of each topic. The automatic topic coherence has been studied extensively, and the normalized point-wise mutual information (NPMI), the probabilistic mutual information (PMI), and the pairwise log-conditional probability (LCP) are co-occurrence metrics that have been shown to correlate positively to the human coherence evaluation \cite{tea}. Thus, we use these metrics to assess the coherence of the extracted topics.

For the experimental settings, we fixed $\alpha$ and $\beta$ to 0.5, and the number of topics $K$ to 25. We varied the value of $N$ to study the effect of defining the coherence with different sets of topics. Finally, we normalized the measurements to be invariant to changes in $N$, so that it facilitates the comparison of the results. %The following equation illustrates the normalization of the LCP metric for the $k^{th}$ topic.

%\vspace{-0.5cm}
%$$
%LCP(k) = \dfrac{2}{(N{-}1){*}(N{-}2)} \sum_{j=2}^{N}\sum_{i=1}^{j-1}log\left(\dfrac{P(w_j,w_i|k)+1}{P(w_i|k)}\right)
%$$

%\vspace{-0.5cm}
%$$
%LCP(k) = \dfrac{2}{(N-1)*(N-2)} \sum_{j=2}^{N}\sum_{i=1}^{j-1}\dfrac{P(w_j,w_i|k)+1}{P(w_j|k)P(w_i|k)}
%$$

\begin{table}
\caption{Automatic topic coherence of LDA inferred by CGS, SAME, and Heron. Best results per dataset use bold font. From top to bottom, the datasets are Cora, Diggs, and Movielens, respectively.}
\label{tab:topiccoherence}
\centering
\setlength\tabcolsep{4pt} 
\begin{tabular}{cc@l^c^c^c^c^c^c}
\toprule
&Method & N &  \multicolumn{2}{c}{LCP} &  \multicolumn{2}{c}{NPMI} &\multicolumn{2}{c}{PMI} \\
&   & mean &  std & mean & std & mean &std \\
\midrule
\multirow{ 6}{*}{\rotatebox{90}{Cora}} &CGS & \multirow{3}{*}{20} &\rowstyle{\cellcolor{hcolor1}} -6.496 & 0.031 & 0.043 & 0.003 & 0.641 & 0.120\\
&SAME &   &\rowstyle{\cellcolor{hcolor2}} -6.083 & 0.024 & \textbf{0.058} & 0.006 & 0.793 & 0.056\\
&Heron &  &\rowstyle{\cellcolor{hcolor3}} \textbf{-6.032} & 0.033 & 0.055 & 0.002 & \textbf{0.823} & 0.070\\
&CGS & \multirow{3}{*}{50} &\rowstyle{\cellcolor{hcolor1}} -6.819 & 0.029 & 0.030 & 0.003 & 0.477 & 0.051\\
&SAME &   &\rowstyle{\cellcolor{hcolor2}} -6.678 & 0.039 & 0.037 & 0.001 & 0.503 & 0.019\\
&Heron &  &\rowstyle{\cellcolor{hcolor3}} -6.619 & 0.021 & 0.030 & 0.002 & 0.548 & 0.021\\
\midrule
\multirow{ 6}{*}{\rotatebox{90}{Diggs}} &CGS & \multirow{3}{*}{20} &\rowstyle{\cellcolor{hcolor1}} -8.693 & 0.001 & 0.046 & 0.001 & 0.941 & 0.002\\
&SAME &   &\rowstyle{\cellcolor{hcolor2}} -8.524 & 0.066 & 0.051 & 0.006 & 1.062 & 0.121\\
&Heron &  &\rowstyle{\cellcolor{hcolor3}} \textbf{-7.483} & 0.040 & \textbf{0.159} & 0.004 & \textbf{3.047} & 0.069\\
&CGS & \multirow{3}{*}{50} &\rowstyle{\cellcolor{hcolor1}} -8.593 & 0.001 & 0.052 & 0.001 & 1.025 & 0.001\\
&SAME &   &\rowstyle{\cellcolor{hcolor2}} -8.176 & 0.024 & 0.085 & 0.002 & 1.681 & 0.004\\
&Heron &  &\rowstyle{\cellcolor{hcolor3}} -8.165 & 0.014 & 0.109 & 0.002 & 2.118 & 0.042\\
\midrule
\multirow{ 6}{*}{\rotatebox{90}{Movielens}} &CGS & \multirow{3}{*}{20} &\rowstyle{\cellcolor{hcolor1}} -6.889 & 0.121 & 0.247 & 0.007 & 4.865 & 0.136\\
&SAME &   &\rowstyle{\cellcolor{hcolor2}} \textbf{-6.527} & 0.000 & \textbf{0.295} & 0.000 & 4.905& 0.000\\
&Heron &  &\rowstyle{\cellcolor{hcolor3}} -6.797 & 0.107 & 0.249 & 0.009 & \textbf{5.838} & 0.178\\
&CGS & \multirow{3}{*}{50} &\rowstyle{\cellcolor{hcolor1}} -6.853 & 0.033 &  0.247 & 0.003 & 4.855 & 0.064\\
&SAME &   &\rowstyle{\cellcolor{hcolor2}} -7.157 & 0.000 &  0.231 & 0.000 & 4.543 & 0.000\\
&Heron &  &\rowstyle{\cellcolor{hcolor3}} -6.812 & 0.029 &  0.250 & 0.002 & 4.921 & 0.043\\

\bottomrule
\end{tabular}
\end{table}

Results presented in Table \ref{tab:topiccoherence} show that Heron is able to infer latent topics with much better or comparable coherence compared to that by CGS or SAME. Interestingly, SAME achieves the best topic coherence in Movielens for $N=20$. Although, simultaneously, it also obtains the worst results for $N=50$. Apparently, SAME assigns a high ranking to the most frequent words in the dataset which leads to an improved coherence as measured by the co-occurrence based metrics.

In all datasets, a smaller value of $N$ results in better coherence. This can be justified by the fact that as we increase the number of words inside a topic, the harder it is to maintain the coherence of the topic. In other words, the top ranked words per topics typically result in better representation for the topic. 

Given that Heron is able to improve CGS in all datasets, we conclude that Heron is able to infer coherent latent topics, answering \textbf{RQ6}.

 %The study of topic coherence for non text data is not deeply studied

\section{Conclusion}
In this paper, we have presented a probabilistic inference method within Gibbs framework named \textit{Heron}. The derivation for Heron begins with maximizing the replications of the latent state, which is known to approximate better local optima. As such, our method achieves a higher predictive power as measured by the predictive perplexity. We studied the properties of convergence of the proposed algorithm and we found that ultimately we are solving a non-linear system of equations. This is beneficial since the fixed-point iterative algorithm that solves the system of equations allow the inference to be distributed into as many processes as the number of observed records in the dataset without affecting the estimation of the parameters. Our proposed Heron is deterministic which reduces the learning process to the pure computation of arithmetic operations and as result, our method runs faster than the state of the art GPU implementation within the Gibbs framework. GPU and CPU implementations of Heron are available at \url{https://github.com/danrugeles/Heron/}.

\vspace{0.3cm}

\bibliographystyle{ACM-Reference-Format}
\bibliography{mybib}

\end{document}